\documentclass[10pt,twocolumn,letterpaper]{article}

\usepackage[accsupp]{axessibility} 
\usepackage{iccv}
\usepackage{times}
\usepackage{epsfig}
\usepackage{graphicx}
\usepackage{amsmath}
\usepackage{amssymb}
\usepackage{subcaption}

\usepackage{authblk}
\usepackage{multirow}

\usepackage[pagebackref=true,breaklinks=true,letterpaper=true,colorlinks,bookmarks=false]{hyperref}

\iccvfinalcopy 

\def\iccvPaperID{8578} 

\ificcvfinal\fi

\graphicspath{{imgs/}}

\makeatletter
\renewcommand\AB@affilsepx{\quad\quad\protect\Affilfont}

\makeatother

\begin{document}

\title{De-rendering Stylized Texts}

\author[1]{Wataru Shimoda}
\author[2]{Daichi Haraguchi}
\author[2]{Seiichi Uchida}
\author[1]{Kota Yamaguchi}
\affil[1]{CyberAgent}
\affil[2]{Kyushu University}

\maketitle
\ificcvfinal\thispagestyle{empty}\fi

\begin{abstract}
Editing raster text is a promising but challenging task.
We propose to apply text vectorization for the task of raster text editing in display media, such as posters, web pages, or advertisements.
In our approach, instead of applying image transformation or generation in the raster domain,
we learn a text vectorization model to parse all the rendering parameters including text, location, size, font, style, effects, and hidden background, then utilize those parameters for reconstruction and any editing task.
Our text vectorization takes advantage of differentiable text rendering to accurately reproduce the input raster text in a resolution-free parametric format.
We show in the experiments that our approach can successfully parse text, styling, and background information in the unified model, and produces artifact-free text editing compared to a raster baseline.
\end{abstract}

\section{Introduction}\label{sec:introduction}
Typography is the art of visually arranging letters and text.
Typography greatly affects how people perceive textual content in graphic design. Graphic designers carefully arrange and stylize text to convey their message in display media, such as posters, web pages, or advertisements.
In computer vision, letters and texts have been predominantly the subject of optical character recognition (OCR), where the goal is to identify characters in the given image.
This includes both OCR in printed documents or automatic scene text recognition for navigation. 
However, often, typographic information of text, such as font, outline, or shadow, is ignored in text recognition because the OCR does not target at reproducing typography in the output.

\begin{figure}[t]
  \begin{center}
  \includegraphics[width=\columnwidth]{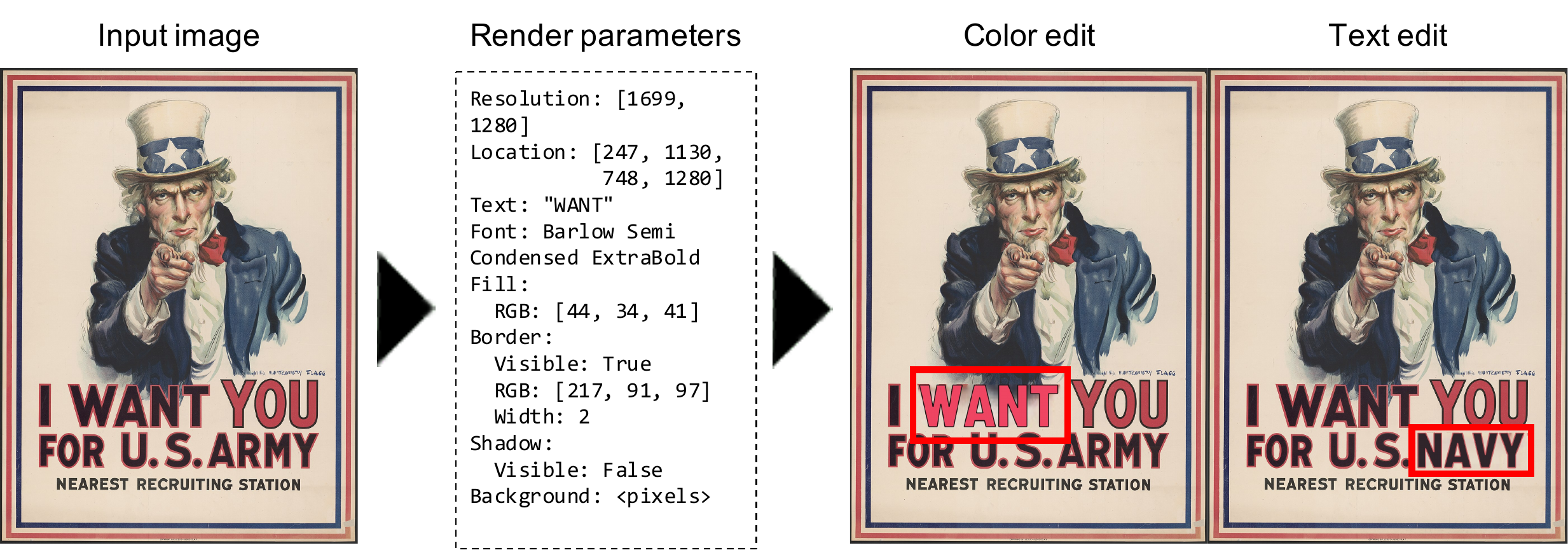}
  \caption{The proposed text editing approach. 
  Once the rendering parameters are recovered, we can apply any text editing and style manipulation.}
  \label{fig:concept}
  \end{center}
\end{figure}

In this paper, we consider raster text editing as a text vectorization problem.
There have been a few attempts at editing raster text mainly in scene images~\cite{stefann,srnet,swaptext}.
Previous work frames text editing as a style transfer problem in the pixel domain, where the goal is to apply the original style and effect in the input image to the target characters in the destination.
However, the major limitation of the style transfer approach is that 1) it is hard to avoid artifacts in the resulting image, and 2) a pixel-based model is resolution-dependent.
Our main target is display media.
Compared to scene text images~\cite{stefann,srnet,swaptext} that often involve external noise due to lighting condition, we often observe noise-free texts in display media which is prone to small artifacts on reconstruction.
Editing in the vector format has a clear advantage in display media, as rendering results in consistent and sharp drawing at any resolution.
Our approach is equivalent to viewing text recognition as a \emph{de-rendering} problem~\cite{neural_derendering}, where the goal is to predict rendering parameters.
Figure \ref{fig:concept} illustrates our approach to the task of raster text editing.

Text vectorization can be an ill-posed problem.
For editing raster text in the vector format, we need to first parse characters, text styling information such as font and effects, and hidden background pixels.
Once those rendering parameters are recovered from the input image, we can edit and render the target text using a rasterizer.
Our approach hence has to solve three sub-problems: 1) OCR, 2) background inpainting, and 3) styling attribute recognition.
OCR has a long history of research that dates back to the 1870s~\cite{schantz1982history}.
In parallel, inpainting has been studied in numerous literature~\cite{inpaint,yu2019free,edgeconnect,liu2018image}.
It might look straightforward to add styling attribute recognition to those two problems for our task.
However, parsing various stylistic attributes is not trivial, as the presence of multiple rendering effects can easily lead to an ill-posed decomposition problem.
For example, it is impossible to decompose the border and fill colors if two colors are the same.
For solving such an ill-posed problem, we train neural networks to predict statistically plausible parameters.

Our model parses text, background, and styling attributes in two steps; we obtain the initial rendering parameters from feedforward inference, then refine the parameters by feedback inference.
Our feedback refinement incorporates differentiable rendering to reproduce the given raster text, and fit the parameters to the raster input using reconstruction loss.
Following SynthText~\cite{synthtext}, we train our feedforward model on text images we generate using a rendering engine.
In the experiments, we show that our vectorization model accurately parses the text information.
Further, we demonstrate that we can successfully edit texts by an off-the-shelf rendering engine using the parsed rendering parameters.

We summarize our contributions below.
\begin{itemize}
\item We formulate raster text editing as a \emph{de-rendering} problem, where the task is to parse potentially ill-posed rendering parameters from the given image.
\item We propose a vectorization model to parse detailed text information, using forward and backward inference procedure that takes advantage of differentiable rendering.
\item We empirically show that our model achieves high quality parsing of rendering parameters. We also demonstrate that the parsed information can be successfully utilized in a 2D graphics engine for downstream edit tasks. We release the codes at Github\footnote{available at \url{https://github.com/CyberAgentAILab/derendering-text}}.
\end{itemize}

\section{Related work}\label{sec:related_work}
\begin{figure*}[t]
  \begin{center}
  \includegraphics[width=\textwidth]{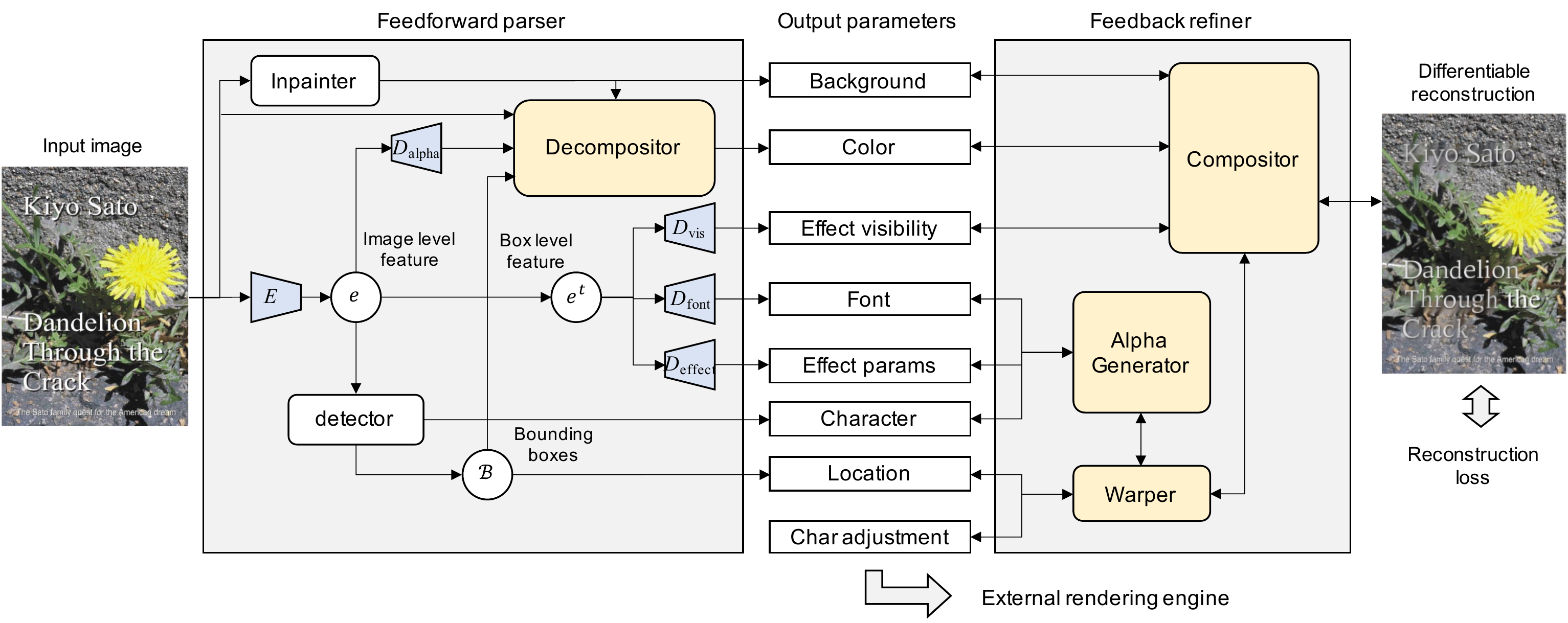}
  \caption{Overview of our approach. Our vectorization model parses rendering parameters from the raster text image in two stages. The feedforward inference gives the initial prediction of rendering parameters, and the feedback inference refines the solution by minimizing reconstruction loss via differentiable rendering.}
  \label{fig:model}
  \end{center}
\end{figure*}

\paragraph*{Optical character recognition}\label{sec:ocr}
OCR has a long history of research.
The earliest ideas of OCR are conceived in the 1870s in Fournier d'Albe's Optophone and Tauschek's reading machine~\cite{schantz1982history}.
In the past, OCR models could not deal with font variations, but recent methods developed for scene texts can detect and recognize any fonts.
Among many literature, EAST~\cite{east} and CRAFT~\cite{craft} are well-known recent detection approaches.
FOTS~\cite{fots} and CharNet~\cite{charnet} report an end-to-end detection and recognition pipelines.
FOTS incorporates a detection model and CRNN after ROI rotation, whereas CharNet utilizes a pixel-wise character classifier for assigning a character label to detect bounding boxes.
In this paper, we adopt the CharNet model for the OCR module, but our model is not restricted to a specific OCR model.

\paragraph{Font recognition}\label{sec:font_recognition}
Compared to OCR, there has been little work for font recognition.
Among a few literature, DeepFont~\cite{deepfont} reports a deep learning approach to categorize fonts.
The major challenge in font recognition is that fonts can be arbitrarily created by font designers, and it is often impossible to define a finite set of font categories.
For this reason, some work instead resorts to font similarity evaluation~\cite{haraguchi_font_cmp,font_att_retrieval}.
In this paper, we only use a fixed set of fonts to avoid the handling of infinite fonts.

\paragraph{Image vectorization}\label{sec:vectorization}
Image vectorization aims at recognizing drawing entities such as lines or shape primitives from an image, and a classic example is line detection by Hough transform.
Image vectorization has been actively studied and is still an open problem~\cite{LineDrawing1,LineDrawing2,TechDrawing,deepsvg,ardeco,gradientmesh,Diffusioncurve,subdivision,semilayers,colortexturevec}.
Jia \etal~\cite{depthordering} propose an approach to preserve an object arrangement and positional relationship using a depth map.
Favreau \etal propose a method for vector clip-arts~\cite{photo2clip}. They decompose the input image into a small number of semi-transparent layers by color gradients.
Liu \etal~\cite{floorplan} report a method to convert a rasterized floor-plan into a vector format. 
Kim \etal\cite{charDrawing} turn Chinese character images into vector formats by converting segmentation to overlapping strokes.
Our approach shares image vectorization approaches in that text effects can be seen as image layers, but the target recognition entity is characters and rendering parameters rather than low-level shape primitives.

\paragraph{Raster text editing}\label{sec:text_editing}
There has been a few recent work on raster text editing based on generative models in the pixel domain.
STEFANN~\cite{stefann} proposes a model that generates a character mask for swapping and transferring color from the source character to the target, then paste the transferred character onto the background.
SRNet~\cite{srnet} takes a raster text image and a target text in pixels, encodes them into a feature space, and generates a stylized target text as a decoding process.
Yang \etal~\cite{swaptext} further improve SRNet by manipulating geometric control points of characters to move text locations.
Our work does not rely on pixel-domain generative models for text and styles but instead predicts parametric representation necessary for reproducing the raster text image.

\paragraph{Scene text inpainting}\label{sec:inpainting}
Apart from editing text, some recent work solely focuses on erasing text by inpainting~\cite{nakamura2017scene,tursun2019mtrnet,zhang2019ensnet,zdenek2020erasing,liu2020erasenet}.
Nakamura \etal~\cite{nakamura2017scene} are one of the first attempts that applies an inpainting approach to erase scene text.
Zdeneck \etal consider text erasing under a weakly supervised setting~\cite{zdenek2020erasing}.
The background inpainting module in our model considers the same problem, yet our model further considers foreground text in terms of rendering.

\paragraph{Text style transfer}\label{sec:text_style_transfer}
Font style transfer is yet another line of relevant work.
Most of the work is based on generative networks except for Awesome Typography~\cite{awesometypography}, where Yang \etal propose to transfer text effects by a patch matching algorithm.
Azadi et al.~\cite{mcgan} train the GAN-based model for character style transfer.
TETGAN~\cite{tetgan} is another GAN-based character style transfer method that preserves various effects.
Yang \etal stylize text with arbitrary texture~\cite{uceffect,iccvtextst}.
Wenjing \etal propose to transfer style for decorative elements~\cite{typographydecor}.
Men \etal transfer video motion to text style~\cite{dyntypo}.

Our work does not consider a highly-complex textured text image.
Instead, we model basic text effects by parameters in the rendering engine including font, fill, border, and shadow, so that we can interpret and easily control the effect for editing.
Although it is our future work to expand the effect kinds, typical typography in display media do not exhibit complex combination of effects.

\section{Our approach}\label{sec:method}
In this paper, we consider raster text editing as a text vectorization problem, where we solve three sub-problems: 1) OCR, 2) background inpainting, and 3) text attribute recognition.
We approach this task by two-stage inference model consisting of a feedforward parser and a feedback refiner.
We illustrate our model architecture in Fig~\ref{fig:model}.

\subsection{Feedforward parsing}\label{sec:vectorization_model}
Our feedforward parser builds upon an OCR model that detects and recognizes characters in the image. We utilize an image-level feature map and box-level feature maps from the OCR model to further parse background colors, alpha maps of effects, and text attributes at each bounding box. 

\paragraph{Detection and recognition}\label{sec:text_detection}
Formally, our model takes a raster image $x$ and first detects a set of bounding boxes $\mathcal{B}(x) = \{b^t, b^c\}$, where $b^t$ and $b^c$ are boundig boxes for texts and characters.
For the detection task, we train an encoder $E$ to obtain a global feature map $e(x) \in {\mathbb R^{C \times H \times W}}$, where $C, H, W$ are the channels, height, width of the feature map.
We use the hourglass-like network~\cite{charnet} to obtain the feature map.
\footnote{We show the detailed architecture in the supplementary.}
The detector localizes the location of characters in the image.
We follow the approach of CharNet~\cite{charnet} and generate a foreground segmentation and a geometry map that predicts distance to the center of the region to obtain localized bounding boxes $\mathcal{B}(x)$.
After localizing text and character regions, we apply a character decoder $D_\mathrm{character}$ to predict a sequence of characters.
Internally, our model combines the foreground segmentation, the geometry map, and a character class map to parse the text in the given region.
Note that we use CharNet~\cite{charnet} to implement our model, but can be any other model as long as characters can be localized.

\paragraph{Font recognition}
Similarly to text, we predict font categories of the given region using the local feature maps at the given bounding box $e_b^t(x)$ and $e_b^c(x)$, which is a subset of the global feature map $e(x)$ at the bounding box $b^t$ and $b^c$. We apply pooling followed by a multi-layer perceptron to classify fonts.

\paragraph{Background inpainting}\label{sec:background_inpainting}
The purpose of the background inpainting is to erase foreground raster text so that afterwards we can render manipulated text on the occluded background region.
We can apply any image inpainting approach, but the challenge here is that we need to erase not only text but also composited text effects.
Since inpainting is a pixel-domain problem, we adopt a pixel-based generator.
Our background inpainter takes a raster text image and predicts background RGB pixels $x_\mathrm{bg}$.
Note that we utilize OCR (Sec~\ref{sec:text_detection}) and pixel-level alpha (Sec~\ref{sec:text_effects}) to specify the region to inpaint.
In our experiment, we adopt either an off-the-shelf inpainting method \cite{telea2004image} or we learn the inpainting model of MEDFE~\cite{inpaint}.

\paragraph{Text effects}\label{sec:text_effects}
Text effects involves a complex compositing process.
We briefly describe the compositing process in the following.
Formally, let us denote an alpha map of the $k$-th effect layer by $\alpha_k(i, j) = f_k(z(i, j), \theta_k)$, where $f_k$ is an effect function that generates an alpha map given the source shape $z$ and $\theta_k$ is the effect parameter.
In our setting, $z$ is a text rasterizer that generates a mask pixel at location $(i, j)$. 
For a plain source-over blending on an opaque background\footnote{Generic blending further considers different color blend modes, but we leave out for simplicity.
See \url{https://www.w3.org/TR/compositing-1/} for further details.}, a rendering engine composites the $k$-th effect layer to the background image in the following equation:
\begin{eqnarray}
\mathbf{c}(i, j) = & (1 - \alpha_k(i, j)) \mathbf{c}_\mathrm{bg}(i, j) + \alpha_k(i, j) \mathbf{c}_k,
\label{eq:compositing}
\end{eqnarray}
where $\mathbf{c}(i, j)$ is the color of the composited image, $\mathbf{c}_\mathrm{bg}(i, j)$ is the background color, and $\mathbf{c}_k$ is the effect color.
When there are multiple effects, a rendering engine repeats the above compositing operation by substituting $\mathbf{c}_\mathrm{bg}(i, j) \leftarrow \mathbf{c}(i, j)$ at each effect $k$.

In this paper, we restrict the available effects to three basic but commonly-used kinds: \emph{fill}, \emph{border}, and \emph{shadow}. 
We also assume the ordering of the effects is fixed to (shadow, fill, border).
In our setting, all effects have color $\mathbf{c}_k$.
Shadow and border effects have visibility parameter $v_k \in \{0, 1\}$ that indicates whether the effect is enabled.
Border has a width parameter: $\theta_\mathrm{border} = \{ p_\mathrm{width} \}$ and shadow has three additional parameters that control the effect generation: $\theta_\mathrm{shadow} = \{ p_\mathrm{blur}, p_\mathrm{dx}, p_\mathrm{dy} \}$.
For each effect $k \in \{ \mathrm{fill}, \mathrm{border}, \mathrm{shadow} \}$, we learn effect decoders from the feature maps:
\begin{eqnarray}
\hat{v}_k &=& D_{v, k}(e_b(x)),\\
\hat{\theta}_k &=& D_{\theta, k}(e_b(x)).
\end{eqnarray}
For decoding color $\mathbf{c}_k$, we first decode the alpha map $\alpha_k$ for the entire canvas, and sequentially decomposite the color following the inverse of eq~\ref{eq:compositing} and averaging the color in the bounding box region:
\begin{eqnarray}
\hat{\alpha}_k &=& D_{\alpha, k}(e(x)),\\
\mathbf{y}_{i,j,k} &=& \frac{\mathbf{c}(i, j) - (1 - \hat{\alpha}_k(i, j))\mathbf{c}_\mathrm{bg}(i ,j)}{\hat{\alpha}_k(i, j)},\\
\hat{\mathbf{c}}_k &=& \arg\max_{\mathbf{y}_{i,j,k}} P(\mathbf{y}_{i,j,k}),
\end{eqnarray}
where the decompositing order is (border, fill, shadow), and if $\hat{v}_k = 0$, we skip decompositing of the effect $k$. 
For the initial border decompositing, we set $\mathbf{c}_b(i, j)$ from the result of the inpainting in Sec \ref{sec:background_inpainting}, and otherwise use the color of the previously decomposited effect.

\subsection{Feedback refinement}\label{sec:refinement}
While the feedforward parsing produces plausible rendering parameters, usually there are still inaccuracy in exactly reproducing the appearance of the input image.
In our feedback refinement stage, we fill the appearance gap between the input and the vector reconstruction by minimizing the error in the rasterized image.
The goal is to minimize the error between the input image $x$ and the raster reconstruction $\mathcal{R}$ with all the rendering parameters $\Theta$:
\begin{eqnarray}
\min_\Theta | \mathcal{R}(\Theta) - x |. \label{eq:refinement_objective}
\end{eqnarray}
The raster reconstruction function $\mathcal{R}$ produces the final reconstruction using compositing equation (eq~\ref{eq:compositing}).
As long as the compositing process is differentiable, we can apply back propagation to efficiently optimize parameters $\Theta$.
We propose to reconstruct a raster image using pre-rendered alpha maps so that the rendering process becomes differentiable.
Note, while there exist recent approaches on differentiable rendering of strokes or shape primitives~\cite{diffvg,beziersketch,kim18semantic}, we are not aware of any existing work on direct vectorization of text styling.

\paragraph{Pre-rendered alpha maps}
In the compositing process (eq~\ref{eq:compositing}), the glyph shape function $z$ and the effect function $f_k$ is usually not differentiable with respect parameters like font category and character class.
We circumvent this issue by replacing the actual effect function by pre-rendering.
The key idea is to rasterize non-differentiable glyphs and effects beforehand so that we can approximate rendering process as a linear combination of pre-rendered shapes.
Let us denote the non-differentiable rendering parameters by $\Theta' = \{\theta_l | l \in \{ \mathrm{char}, \mathrm{font}, \cdots \}\} \subset \Theta$.
We approximate the non-differentiable shape function $z$ by:
\begin{align}
    z(i, j; \Theta') &= \sum_{\Theta'} P(\Theta') A_{\Theta'}(i, j), \label{eq:pre-rendering}
\end{align}
where $P(\Theta') = \prod_l P(\theta_l)$ is the joint probability (or so-called attention map) of parameter $\Theta'$, and $A_{\Theta'}(i, j)$ is the pre-rendered alpha maps for the corresponding parameter set $\Theta'$.
Since $\Theta'$ is defined over a finite set, we can pre-render $A_{\Theta'}(i, j)$ for all possible combination of $\Theta'$.
$P(\Theta')$ is differentiable with respect to $\Theta'$, and consequently eq~\ref{eq:pre-rendering} becomes also differentiable.
Similarly, we approximate the borderline effect function $f_k$ in the same pre-rendering approach to make differentiable approximation.
An alpha map constructed from eq~\ref{eq:pre-rendering} visually looks blurred because the shape is a linear combination of different font glyphs.
However, the purpose of this alpha construction is to compute gradients over $\Theta'$ and not to produce the final raster reconstruction.
 
In this work, we use 100 fonts for the pre-rendered alpha maps randomly picked from Google Fonts\footnote{\url{https://github.com/google/fonts}}. 
Note that the pre-rendered alpha maps require large memory footprint if the parameter space is huge; e.g., if there are 100 fonts for 100 characters, we have to pre-render and keep $100 \times 100$ alpha maps, and also computing shape $z$ becomes expensive.
In practice, we compute sparse attention map $P(\Theta')$ for only the top 20 fonts and zero-out the rest to save computation.
We choose $64 \times 64$ resolution for alpha maps considering the balance between the reconstruction quality and the memory footprint.

\paragraph{Character adjustment}
Although the OCR model gives the initial prediction of character bounding boxes, those locations do not perfectly align with the actual position of each character in the given input image.
We introduce image warping to adjust the location of each character during feedback refinement.
We prepare affine transformation parameters for each character and also for the word-level bounding box, and solve for the best adjustment during the optimization (eq~\ref{eq:refinement_objective}).
We set initial warping parameters to identity transformation at the beginning.

\paragraph{Differentiable compositing}
We use pre-rendered alpha for fill and border effects, but shadow has differentiable effect such as blur or offset.
We directly implement those differentiable operators for shadow effect $f_\mathrm{shadow}$.

To reconstruct an image, we composite alpha maps by eq~\ref{eq:compositing} and image warping in character adjustment.
The visibility flag controls whether to enable the effect compositing.
All the operators are differentiable with respect to parameters $\Theta$.
We binalize the effect visibility $v$ with differentiable binarization~\cite{db} and multiply to the foreground alpha.
Note that our differentiable reconstruction is only for refinement purpose.
The final output of our model is the refined rendering parameters $\Theta$.

\begin{figure}[t]
  \begin{center}
  \includegraphics[width=\columnwidth]{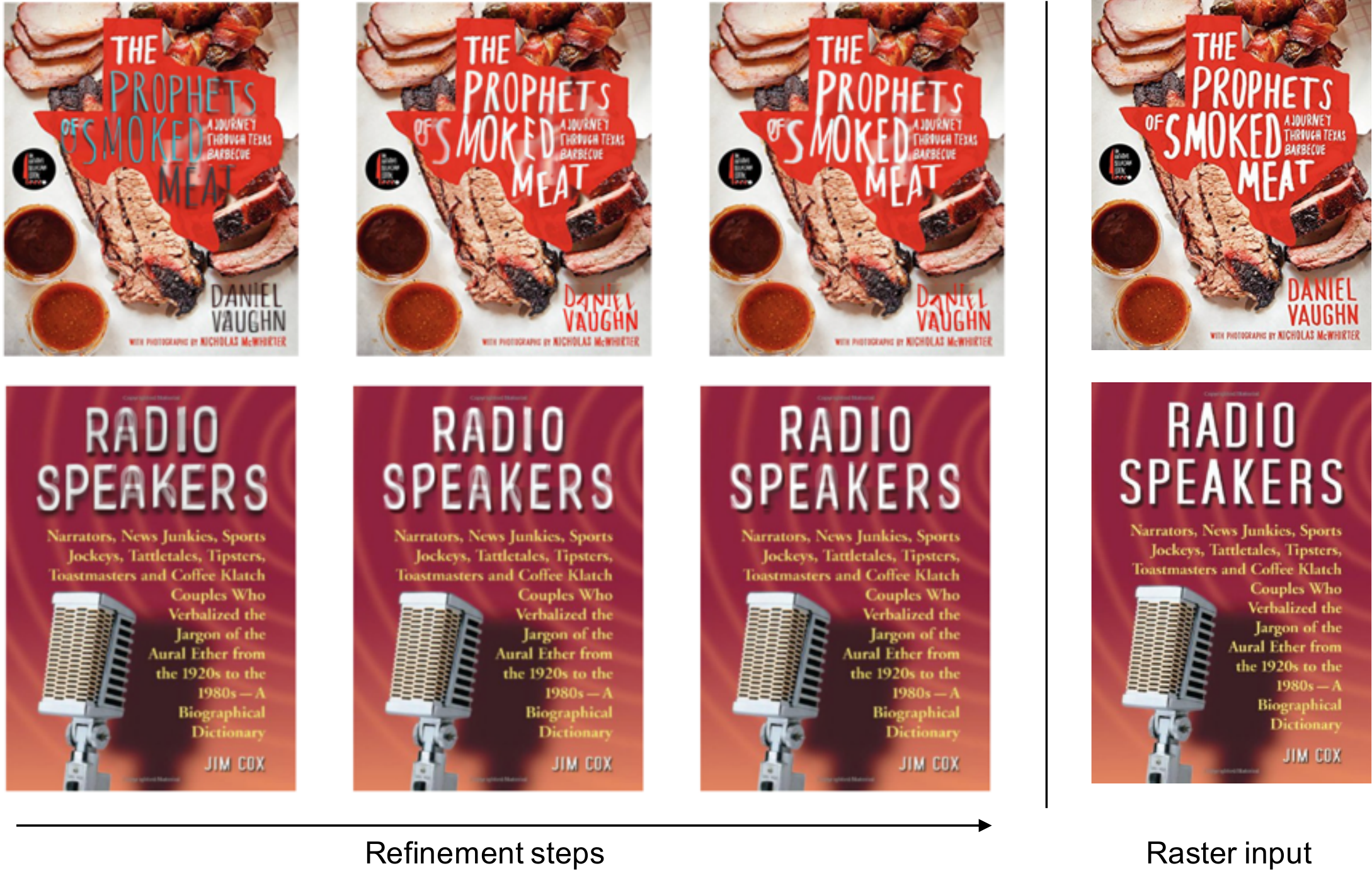}
  \caption{Intermediate outputs during feedback refinement.}
  \label{fig:optimization}
  \end{center}
\end{figure}

We show in Fig~\ref{fig:optimization} the intermediate results during feedback refinement.
Our refinement process starts from the rough initialization by feedforward parser, and gradually fits the parameters to reproduce the given raster image.

\paragraph{Refinement details}\label{sec:refinement_details}
We do not optimize location and background during refinement, because bounding box generation in our OCR model is not differentiable.
Background pixels are differentiable, but we observe unstable behavior during optimization, perhaps due to ill-posed decompositing setup.
We also do not optimize character probability maps of OCR because of unstable behavior.
We refine parameters using Adam optimizer with 200 iterations.
In our implementation, optimization requires about 1 minute for one word.

\subsection{Exporting rendering parameters} \label{sec:export}
Parsed rendering parameters can be directly used in external 2D graphics rendering engine, though we need to adjust the scale of each parameter for the external format.
Most of the parameters can be exported to the rendering engine with simple normalization, but we need to take care of computing geometric information such as font size or offset that our parser does not predict.
Although our OCR model produces bounding boxes, it is not trivial to convert to those geometric information.

We convert these information in the following post-processing.
We first transform character bounding boxes using affine parameters we obtain in letter adjustment.
Next, we obtain a word-level bounding box by taking the minimum and maximum coordinates from the OCR bounding box and the set of character bounding boxes.
Once we obtain both the character- and word-level bounding boxes, we use them to as guidance and conduct a grid search over configuration, starting from the parameters of best-fit character in the refinement.

\section{Training}\label{sec:training}
\subsection{Training objective}\label{sec:training_objective}
For training our model, we minimize the following loss function, where each term corresponds to training of each decoder module in our model:
\begin{eqnarray}
\mathcal{L} &=& \mathcal{L}_{\rm ocr} + \mathcal{L}_{\rm inpaint} + \lambda_{\rm font}\mathcal{L}_{\rm font} +\\ \nonumber
& & \sum_k \lambda_{\rm alpha}\mathcal{L}^k_{\rm alpha} + \mathcal{L}^k_{\rm visibility} + \mathcal{L}^k_{\rm param}.
\end{eqnarray}
The loss functions for text recognition ${\mathcal L}_{\rm ocr}$ and background inpainting $\mathcal{L}_{\rm inpaint}$ are identical to those defined in CharNet~\cite{charnet} and MDEDF~\cite{inpaint}. $\mathcal{L}_{\rm font}$ is the categorical softmax cross entropy for font categories. $\mathcal{L}^k_{\rm alpha}$ is the mean squared error of pixel-wise difference of the effect alpha map. $\mathcal{L}^k_{\rm visibility}$ is the binary cross entropy of the effect visibility. $\mathcal{L}^k_{\rm param}$ is the total of mean squared error for effect parameters. In the experiment, we apply weighting to each term for better convergence.
$\lambda_{\rm font}$ and $\lambda_{\rm alpha}$ are the hyperparameters for balancing. We empirically set the hyperparameter values to $\lambda_{\rm font}=0.1$ and $\lambda_{\rm alpha}=10$. 

\begin{figure}[t]
  \begin{center}
  \includegraphics[width=\columnwidth]{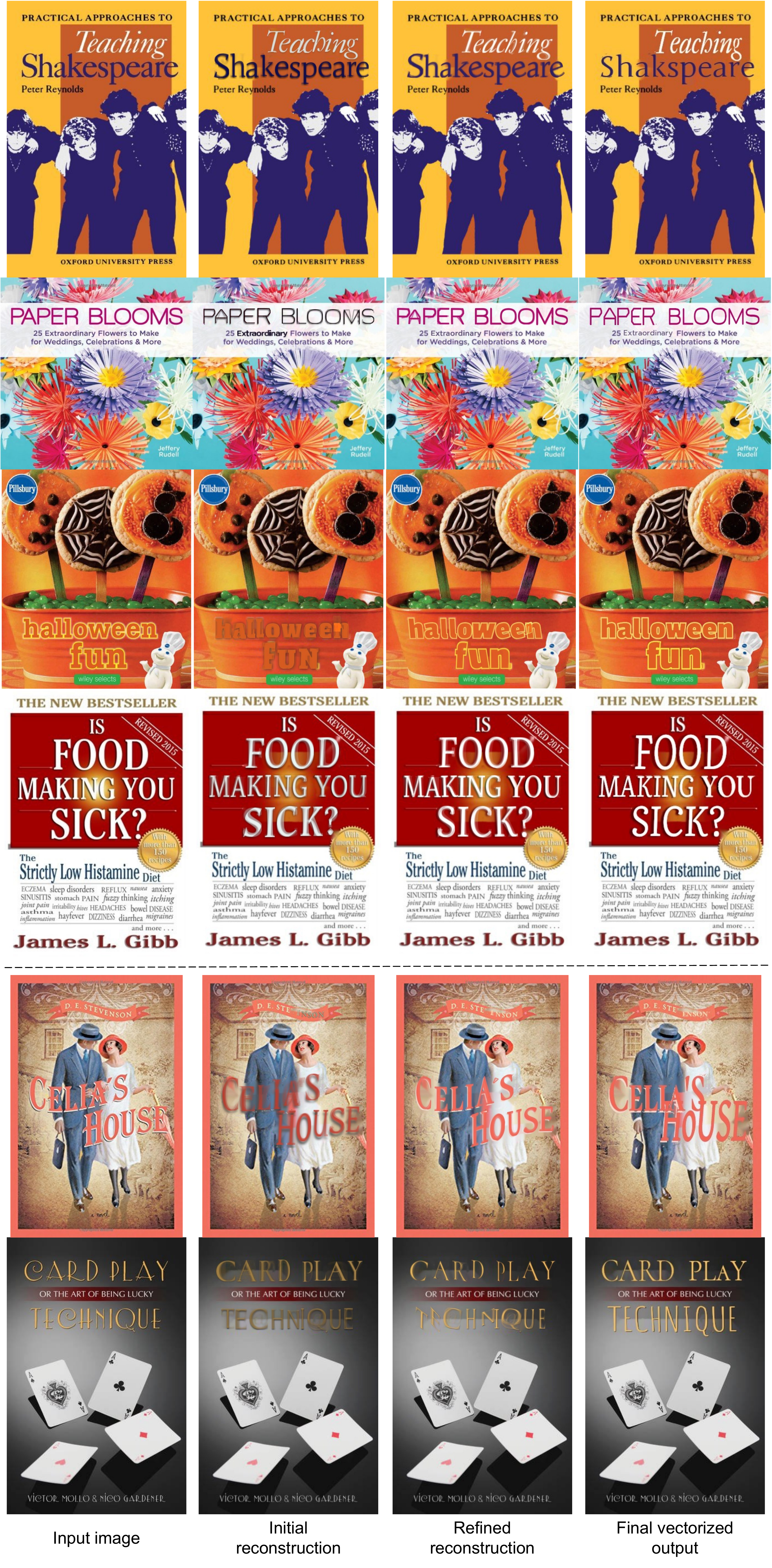}
  \caption{Our two-stage vectorization process. From left to right, we show input images, reconstructions from the initial feedforward prediction, differentiable reconstructions of the feedback refinement, and the final vectorized images rendered using external graphics engine.}
  \label{fig:reconstruction}
  \end{center}
\end{figure}

\subsection{Training data generator}\label{sec:data_generator}
We follow the approach of SynthText~\cite{synthtext} to generate text images for training. 
In addition to the scene background images from SynthText, we collect background images from single-color background data, book cover dataset~\cite{iwana2016judging}, BAM dataset~\cite{bam} and FMD dataset~\cite{fmd}.
For the book cover dataset and BAM dataset, we erase text regions by text detection and inpainting, then generate candidate locations from a saliency map.
We take the vocabulary set from SynthText.
We implement our data generator using the Skia graphics library\footnote{\url{https://skia.org/}}, which can handle both font and effects in a purely resolution-free manner.

\subsection{Implementation detail}\label{sec:implementation_detail}
We pre-train the OCR model for 5 epochs with ~80,000 SynthText images. The condition is the same as the pre-training of CharNet~\cite{charnet}.
The resolution of the input image is $640 \times 640$ , and the feature map is $160 \times 160$ with the channel size to be $256$.
We train our model with the SGD optimizer.
We set the learning rate to $10^{-3}$ to the pre-trained encoder and $10^{-2}$ to all the other decoders, with batch size 4.
We decrease the learning rate with cosine warm up~\cite{cosign_warm}.

We independently train the inpainting model on the generated text data~\footnote{We also attempted joint training in the experiment but did not observe a clear advantage.}.
In the training phase, we make holes for text regions with a mask generated by thresholding alpha values in each effect, and train to fill the holes.
At test time, we make holes for text regions with predicted alpha maps, then fill the holes using the inpainting model.
The resolution of the inpainting output~\cite{inpaint} is $256 \times 256$.
We resize the inpainted image to the target image size.
This upsampling often causes undesired artifacts.
To alleviate the quality issue, we replace text regions with text erased images and blend using predicted alpha.

\begin{figure*}[t]
  \begin{center}
  \includegraphics[width=\textwidth]{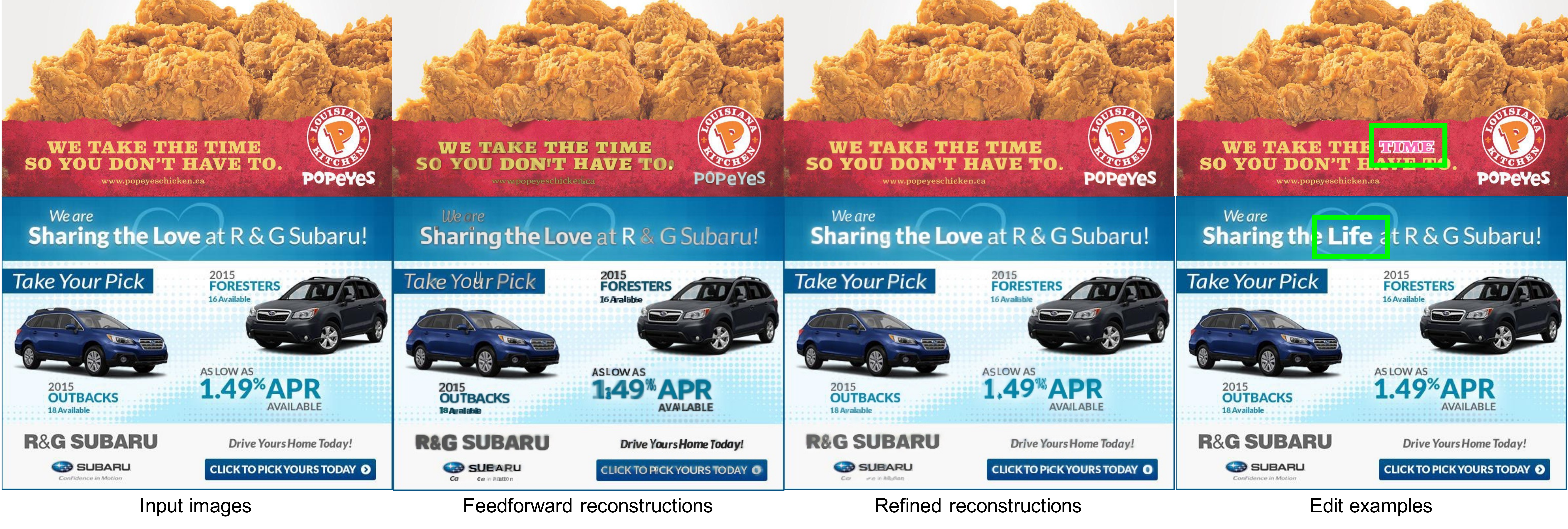}
  \caption{Reconstruction and text editing examples from advertisement dataset. The leftmost column is the input, the second and the third columns are our differentiable reconstruction, and the last column is reconstructed and edited examples from the external graphic engine.}
  \label{fig:ad_data}
  \end{center}
\end{figure*}

\section{Experiments}\label{sec:experiments}
In this experiments, we evaluate 1) how accurately our model parse text information in vectorized representation, and 2) how well our model can perform raster text editing through vectorized information.
For experiments, we use two datasets, the book cover dataset~\cite{iwana2016judging}
and the advertising image dataset~\cite{ad_dataset}.

\begin{figure*}[t]
  \begin{center}
  \includegraphics[width=\textwidth]{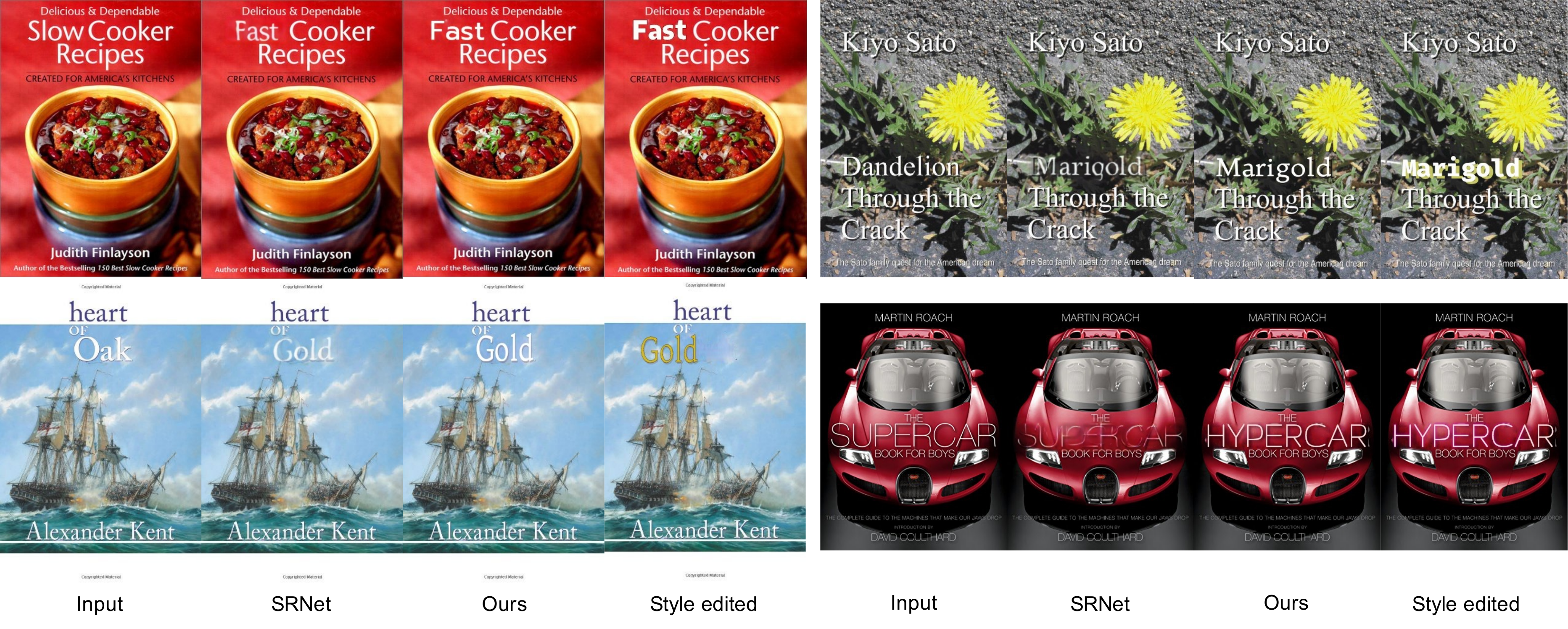}
  \caption{Text and style editing examples.}
  \label{fig:cmp}
  \end{center}
\end{figure*}

\subsection{Qualitative results}
We show in 
Fig~\ref{fig:reconstruction} input images, initial reconstructions after feedforward parsing, reconstructions after feedback refinement, and final vectorization results using a 2D graphics engine from the book cover dataset.
We observe that our approach successfully parses text information in various situations, including stylized texts with effects.
In Fig~\ref{fig:reconstruction}, the first and second rows show examples of simple texts, and the third to fourth rows show examples of stylized texts with thin borderline and small shadow.
The results demonstrate that our model can parse detailed effects parameters such as the weight of borderline and shadow orientation.
The last two rows show challenging examples; the fifth row is an example of non-aligned text, and the last row is an example of unusual font.
Our model parses non-aligned texts, but the vector reconstruction fails to render with the same style.
When there is an unusual font, even though our reconstruction is inaccurate, our feedback refinement tries hard to adjust the choice of font and affine warping parameters to visually match the result.

We show reconstruction and text editing examples in advertisement datasets in Fig~\ref{fig:ad_data}.
Compared to book covers, advertisement images have various aspect ratios.
Our approach successfully reconstructs complex text layouts in vector format with correct styling.

\paragraph{State of the art comparison} \label{sec:sota_comparison}
We compare the quality of our editing approach with a state-of-the-art raster editing method SRNet~\cite{srnet}.
SRNet takes a source style image and a target text mask to render the stylized target, where SRNet converts the text mask into a skeleton appearance and generates an image by transferring style from the reference style source.
We use the alpha mask generated by the external rendering engine for the target text.
Fig~\ref{fig:cmp} shows qualitative comparisons between SRNet and our vectorization approach.
We observe that SRNet often fails to draw a straight line and prone to be small artifacts, while our approach completely separates text shape from text effects in the parametric representation and does not suffer from artifacts.
Note that our result has an additional benefit of easy and reproducible editing in the rendering parameters.

\subsection{Quantitative evaluation}

\begin{table}[t]
  \centering
  \caption{Reconstruction performance.} \label{tab:performance}
  \begin{tabular}{|l|l|cc|}
    \hline
    Dataset & Method  & L1 $\downarrow$ & PSNR $\uparrow$\\
    \hline
    \multirow{3}{*}{Book cover} & Feedforward & 0.15 & 29.22  \\
    & w/ refinement & 0.08 & 30.02 \\
    & External rasterizer & 0.18 & 29.73  \\
    \hline
    \multirow{3}{*}{Advertisement} & Feedforward & 0.16 & 28.77  \\
    & w/ refinement & 0.09 & 29.37 \\
    & External rasterizer & 0.19 & 29.16 \\
    \hline
  \end{tabular}
\end{table}

\begin{table}[t]
\centering
\caption{Component contribution in the book cover dataset.}\label{tab:psnr}
\begin{tabular}{|l|c|}
    \hline
    Model & PSNR$\uparrow$ \\
    \hline
    Full refinement & 30.02 \\
    \hline
    w/o Color & 29.53 \\
    w/o Font & 29.78 \\
    w/o Shadow & 29.79 \\
    w/o Border & 29.99 \\
    \hline
    No refinement & 29.22 \\
    \hline
\end{tabular}
\end{table}

We quantitatively evaluate our model in terms of the reconstruction quality.
We set up the following benchmark conditions.

\noindent\textbf{Book cover} We randomly pick 1,000 images from the book cover dataset, and apply an off-the-shelf OCR model~\cite{craft} to extract title text region of the image for evaluation.
In the book cover dataset, we use Telea's inpainting method~\cite{telea2004image} in our feedforward parser, because we find the off-the-shelf method already works well for book cover images.

\noindent\textbf{Advertisement} We randomly pick 508 images from the ad dataset and evaluate reconstruction performance on all text regions. We use the learned inpainting model~\cite{inpaint}.

To measure the reconstruction performance, we use the average L1 loss and peak signal-to-noise ratio (PSNR) between the input image and the reconstructed image.
We compare the quality of 1) the differentiable reconstruction after feedforward parsing, 2) after feedback refinement, and 3) the final reconstruction in the external graphic engine.

We summarize the evaluation results in Table~\ref{tab:performance}.
We find that our two-stage vectorization approach effectively improves the reconstruction quality, as the refinement consistently improves both L1 and PSNR compared to the feedforward parser only.
The final reconstruction using the external rasterizer does not necessarily gives the identical result to our differentiable reconstruction.
Our refinement process includes approximation and also the exporting process (Sec~\ref{sec:export}) can introduce errors.
Nevertheless, we find the reconstruction quality reasonably well, and PSNR is consistently better than the initial feedforward reconstruction.
Filling the gap between the differentiable reconstruction and the exported result is our future work.

\paragraph{Component analysis}
We show in Table~\ref{tab:psnr} how PSNR changes as we skip feedback refinement in rendering parameters in the book cover dataset.
We skip \emph{color}, \emph{font}, \emph{shadow}, and \emph{border} effects in this ablation study.
As the result suggest, effect components that has larger area in the resulting appearance have more performance influence on the reconstruction quality; incorrect \emph{color} most negatively affect the resulting visual quality, while other effects occupy smaller regions and therefore quantitatively have less minor impact on PSNR.

\subsection{Discussion}
One drawback of our approach is that we cannot handle arbitrary styles that are not defined in our basic effect models or unusual glyph shapes.
We also do not consider detailed rendering parameters such as kerning or geometric transformation.
It is our future work to extend the available fonts or effects, where we have to address efficient differentiable rendering for a large combination of non-differentiable operations (Sec~\ref{sec:refinement}).
However, we note that our basic effects already cover the majority of real-world typography in display media.

\section{Conclusion}
We present a vectorization approach to edit raster text images.
Our approach inpaints the background and accurately parses foreground text information for reconstruction in an external rendering engine.
We show in the experiments our model successfully reproduces the given raster text in vector graphic format.
Our vectorization approach has a clear advantage in resolution-free editing, especially for display media.
One limitation of our approach is that effects must be pre-defined by the rendering engine.
We wish to expand the available effects and blending operations in the future.

{\small
\bibliographystyle{ieee_fullname}
\bibliography{main}
}

\cleardoublepage
\section{Supplementary material}

\subsection{Effect parameter details}
\begin{figure}[b]
\begin{subfigure}[b]{.49\linewidth}
\centering
\includegraphics[width=\columnwidth]{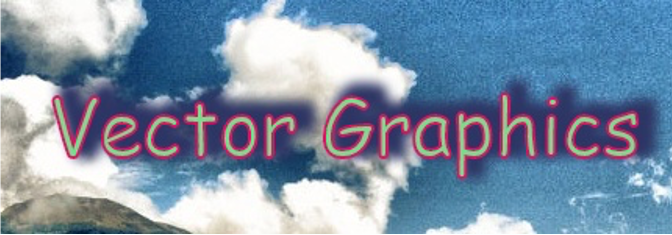}
\caption{Composited}\label{fig:effect_image}
\end{subfigure}
\hfill
\begin{subfigure}[b]{.49\linewidth}
\centering
\includegraphics[width=\columnwidth]{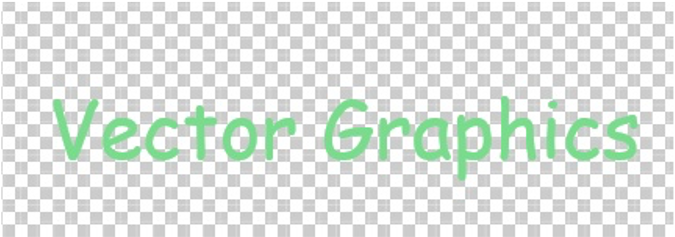}
\caption{Fill}\label{fig:effect_fill}
\end{subfigure}
\begin{subfigure}[b]{.49\linewidth}
\centering
\includegraphics[width=\columnwidth]{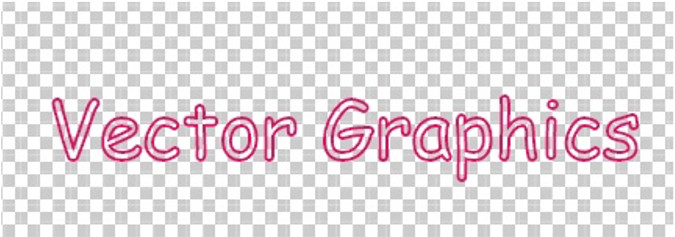}
\caption{Border}\label{fig:effect_border}
\end{subfigure}
\hfill
\begin{subfigure}[b]{.49\linewidth}
\centering
\includegraphics[width=\columnwidth]{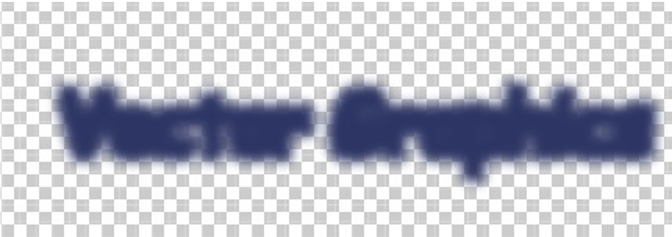}
\caption{Shadow}\label{fig:effect_fill}
\end{subfigure}
\caption{Effect examples.}\label{fig:effect}
\end{figure}

In the experiment, we assumed the most common three text effects: \emph{fill}, \emph{border}, and \emph{shadow}. Fig~\ref{fig:effect} illustrates an example of those effects.
We also assume the ordering of the effects is fixed to shadow $\to$ fill $\to$ border.

\subsection{Illustration of alpha generation}
\begin{figure*}[tb]
  \centering
  \includegraphics[width=0.9\textwidth]{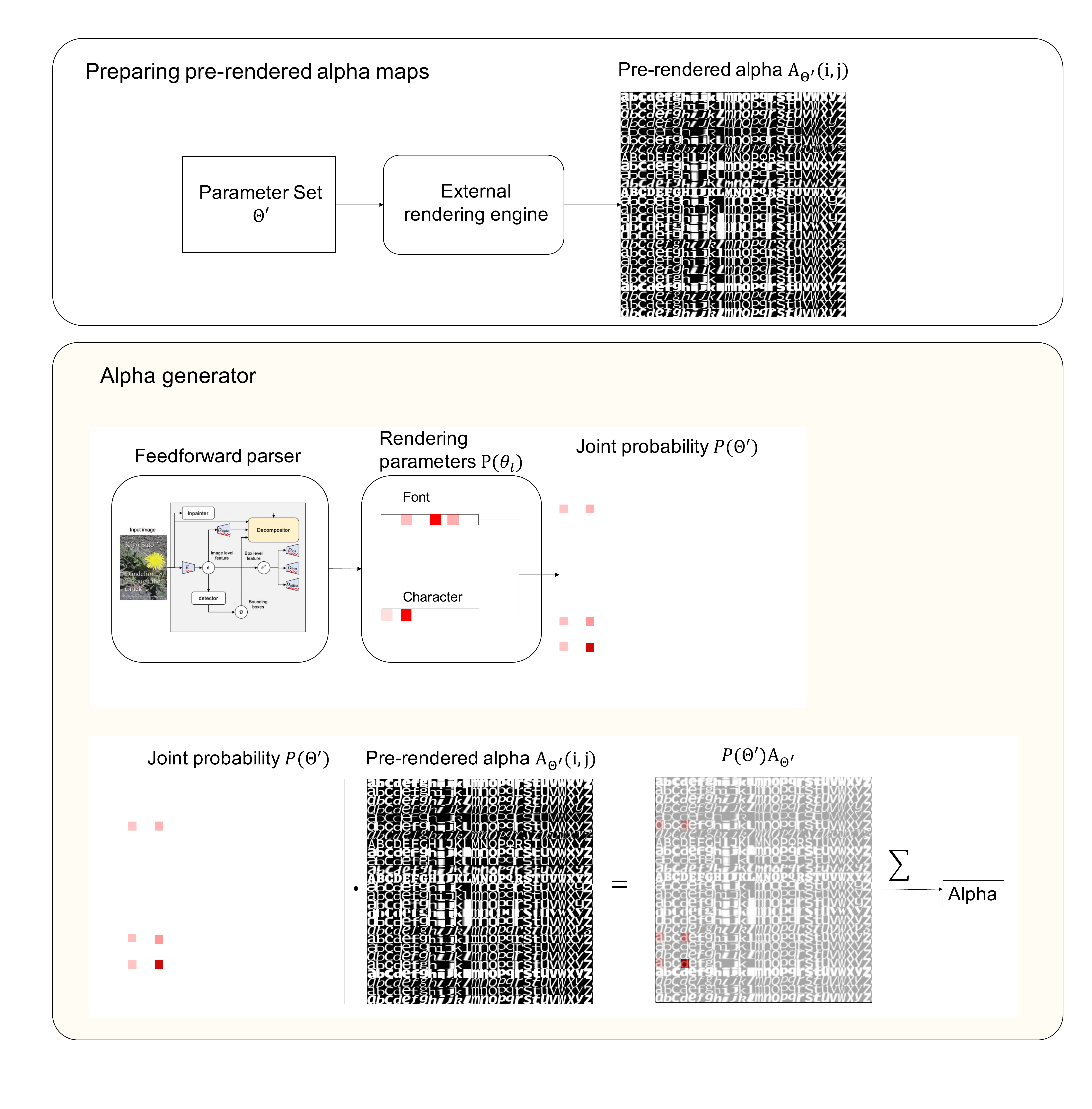}
  \caption{Alpha generator using pre-rendered alpha maps. We illustrate the case where we have 26 characters and fonts.}
  \label{fig:pre-rendered_alpha}
\end{figure*}

We illustrate the process of alpha generation based on pre-rendered alpha maps in Fig~\ref{fig:pre-rendered_alpha}.

\subsection{Text style transfer examples}
As shown in Fig.6, the proposed model can use the text style of another text on the rendering step. Here, we show more detailed examples of text style transfer, as Fig.7. Five different styles are transferred to a text (e.g., ``FREE'' and ``BACON'') in an input image. Note again that the proposed model can transfer not just font style but also effect style.

\subsection{Text image generator details}
We use a text image generator, which is modified from SynthText, and record the exact rendering parameter used in the text image generator to supervise the training of our model.
Here, we introduce our text image generator details, in terms of the type of the background images, text placement rule, and sampling rule of the rendering parameter.

The background images originally used in SynthText are insufficient for the robust performance on the display media; we therefore add background images from single-color background data, Book cover dataset, BAM dataset, and FMD dataset.
For those five types of background images, we render texts and place texts in background images.

The rule of the text placement procedure is different in the type of the background images, respectively.
For the background images from SynthText, the text placement procedure follows SynthText; we apply over-segmentation to generate candidate regions to place a text.
Note that we disable text rotation for the function of SynthText.
For BAM and Book cover data, we utilize a saliency map to generate candidate locations.
For Book cover data, we also generate candidate locations by applying the existing OCR model.
Unlike other background images, BAM and Book cover images often include texts, then we apply text inpainting to erase texts in advance.
We randomly generate text locations for single-color flat backgrounds and FMD images.
On FMD, we crop material regions and use those regions as backgrounds.
We generate 10,000 text rendered images for the background images, respectively.
Then we exclude images that have too-small candidate regions for locating texts, and finally obtained a total of 42,285 images.

After generating candidate regions, we set rendering parameters.
Font categories are randomly sampled from predefined categories.
The effect parameters and colors are randomly sampled from a predefined range.
Unlike SynthText, we implement our data generator using the Skia graphics library\footnote{\url{https://skia.org/}}, which can handle both font and effects without raster artifacts.

\begin{figure*}[tb]
  \centering
  \includegraphics[width=1.0\textwidth]{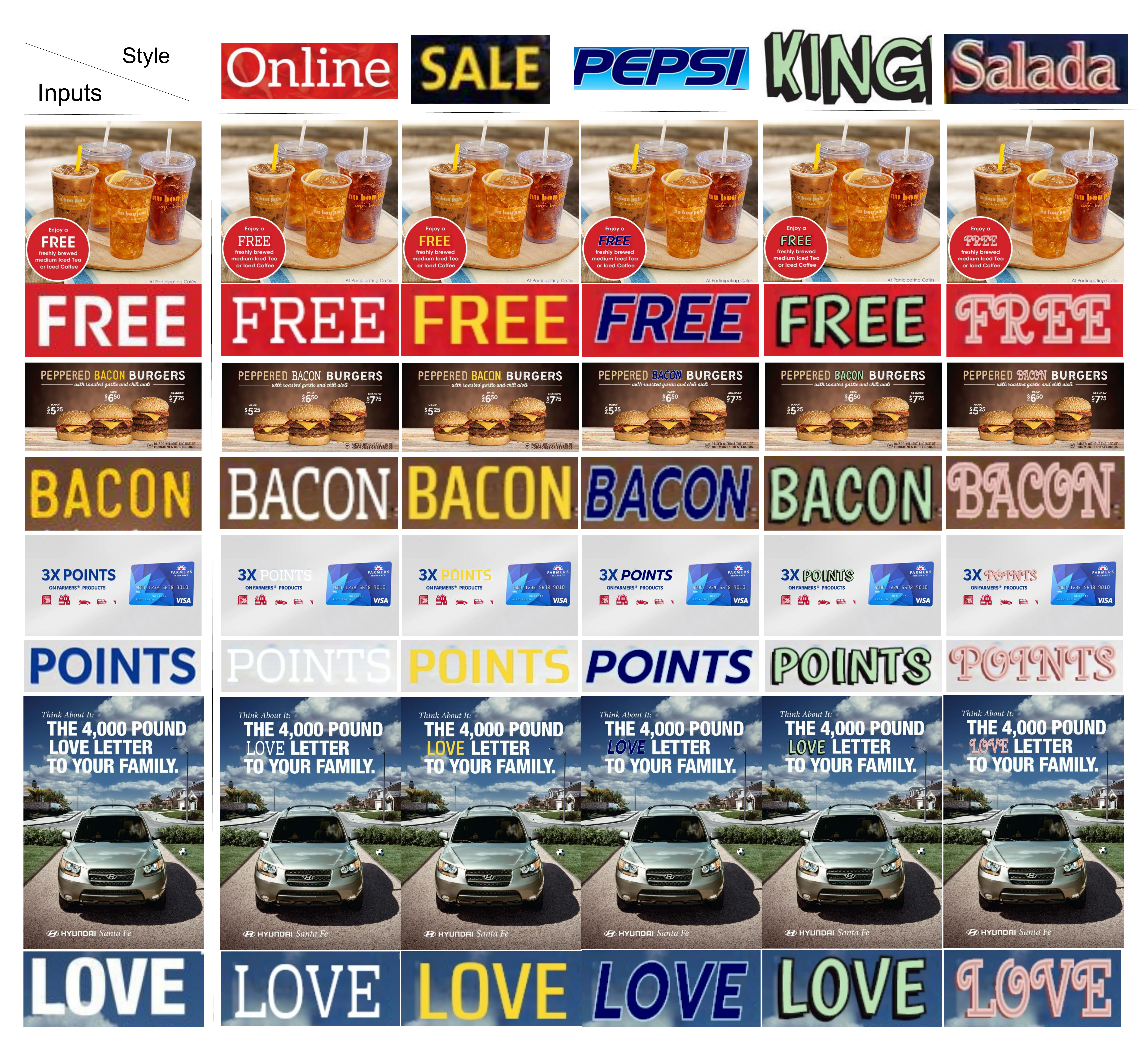}
  \caption{Examples of text style transfer in an external renderer. Since we transfer style information in parameter space, we produce no pixel artifacts such as blur on texts.}
  \label{fig:style}
\end{figure*}

\subsection{Architecture details}
We show the detailed configurations for the parser models in Table~\ref{tab:config}.
Our encoder model, i.e. backbone model, is based on an hour-glass model.
We add some convolution layers to the outputs' head to enlarge the receptive fields because texts tend to be large in the display media.
There are branches for predicting text rendering parameters: the OCR branch, the alpha branch, the font branch, the effect visibility brach, and the effect parameter branch. 
The OCR branch is further split into the word detector, the character detector, and the character classifier.
For extracting text colors, we predict pixel-wise alpha maps to decompose an image.
We obtain font information for each text by a classification model.
To parse text effects, we predict both effects visibility and effects parameters.
We consider visibility prediction as a binary classification problem and shadow parameter estimation as a regression problem.
We predict discretized parameters by a classification model for border effects because we use pre-rendered alpha maps for them.
We quantize the border parameter into five bins in our experiments.

In Table\ref{tab:config}, the third ``kernel and stride'' column indicate the kernel size and the stride in the convolution layer if the layer has those configurations.
The three numerical values in both columns of input size and output size represent tensors' size for channel, height, and width.
We represent the intermediate representation in inputs and outputs by B1-B5 for the backbone model, W1-W3 for the word detector, C1-C3 for the character detector, R1-R3 for the character classifier, A1-A6 for the alpha model, F1 for the font model, V1 for the effects visibility model, and P1 for the effects parameter model.

\begin{table*}[b]
\centering
\caption{Architecture details. }\label{tab:config}
\scalebox{0.8}{
\begin{tabular}{|c|c|c|c|c|c|c|}
    \hline
    Model & Layers & Kernel, Stride &Input & Input size & Output & Output Size \\
    \hline
    \multirow{6}{*}{Backbone} & Hour Glass Net & - & $x$ & $3 \times \rm{H} \times \rm{W}$ &B1 & $256 \times \rm{H}/4 \times \rm{W}/4$ \\
     & CONV + BN + RELU & (3 $\times$ 3), (2 $\times$ 2) & B1 & $256 \times \rm{H}/4 \times \rm{W}/4$ & B2 & $128 \times \rm{H}/8 \times \rm{W}/8$ \\
     & CONV + BN + RELU  & (3 $\times$ 3), (2 $\times$ 2)& B2 & $128 \times \rm{H}/8 \times \rm{W}/8$ & B3 & $128 \times \rm{H}/16 \times \rm{W}/16$ \\
     & CONV + BN + RELU &  (3 $\times$ 3), (2 $\times$ 2) & B3 & $128 \times \rm{H}/16 \times \rm{W}/16$ & B4 & $128 \times \rm{H}/32 \times \rm{W}/32$ \\
     & Upsampling &  - & B4 & $128 \times \rm{H}/32 \times \rm{W}/32$ & B5 & $128 \times \rm{H}/4 \times \rm{W}/4$ \\
     & CONV + BN + RELU &  (3 $\times$ 3), (1 $\times$ 1) & B1 and B5 & $384 \times \rm{H}/4 \times \rm{W}/4$ & $e(x)$ & $256 \times \rm{H}/4 \times \rm{W}/4$ \\
    \hline
    \multirow{5}{*}{Word detector} & CONV + BN + RELU & (3 $\times$ 3), (1 $\times$ 1) & $e(x)$ & $256 \times \rm{H}/4 \times \rm{W}/4$ & W1 & $128 \times \rm{H}/4 \times \rm{W}/4$ \\
     & CONV + BN + RELU & (3 $\times$ 3), (1 $\times$ 1) & W1 & $128 \times \rm{H}/4 \times \rm{W}/4$ & W2 & $32 \times \rm{H}/4 \times \rm{W}/4$ \\
     & CONV + BN + RELU & (3 $\times$ 3), (1 $\times$ 1) & W1 & $128 \times \rm{H}/4 \times \rm{W}/4$ & W3 & $32 \times \rm{H}/4 \times \rm{W}/4$ \\
     & CONV & (1 $\times$ 1), (1 $\times$ 1) & W2 & $128 \times \rm{H}/4 \times \rm{W}/4$ & Text foreground map & $2 \times \rm{H}/4 \times \rm{W}/4$ \\
     & CONV + RELU & (1 $\times$ 1), (1 $\times$ 1) & W3 & $128 \times \rm{H}/4 \times \rm{W}/4$ & Text geometry map & $32 \times \rm{H}/4 \times \rm{W}/4$ \\
    \hline
    \multirow{5}{*}{Char detector} & CONV + BN + RELU & (3 $\times$ 3), (1 $\times$ 1) & $e(x)$ & $256 \times \rm{H}/4 \times \rm{W}/4$ & C1 & $128 \times \rm{H}/4 \times \rm{W}/4$ \\
     & CONV + BN + RELU & (3 $\times$ 3), (1 $\times$ 1) & C1 & $128 \times \rm{H}/4 \times \rm{W}/4$ & C2 & $32 \times \rm{H}/4 \times \rm{W}/4$ \\
     & CONV + BN + RELU & (3 $\times$ 3), (1 $\times$ 1) & C1 & $128 \times \rm{H}/4 \times \rm{W}/4$ & C3 & $32 \times \rm{H}/4 \times \rm{W}/4$ \\
     & CONV  & (1 $\times$ 1), (1 $\times$ 1) & C2 & $128 \times \rm{H}/4 \times \rm{W}/4$ & Char foreground map & $32 \times \rm{H}/4 \times \rm{W}/4$ \\
     & CONV + RELU & (1 $\times$ 1), (1 $\times$ 1) & C3 & $128 \times \rm{H}/4 \times \rm{W}/4$ & Char geometry map & $32 \times \rm{H}/4 \times \rm{W}/4$ \\
    \hline
    \multirow{4}{*}{Char recognizer} & CONV + BN + RELU & (3 $\times$ 3), (1 $\times$ 1) & $e(x)$ & $256 \times \rm{H}/4 \times \rm{W}/4$ & R1 & $128 \times \rm{H}/4 \times \rm{W}/4$ \\
     & CONV + BN + RELU & (3 $\times$ 3), (1 $\times$ 1) & R1 & $128 \times \rm{H}/4 \times \rm{W}/4$ & R2 & $32 \times \rm{H}/4 \times \rm{W}/4$ \\
     & CONV + BN + RELU & (3 $\times$ 3), (1 $\times$ 1) & R2 & $128 \times \rm{H}/4 \times \rm{W}/4$ & R3 & $32 \times \rm{H}/4 \times \rm{W}/4$ \\
     & CONV & (1 $\times$ 1), (1 $\times$ 1) & R3 & $94 \times \rm{H}/4 \times \rm{W}/4$ & Char recognition map & $32 \times \rm{H}/4 \times \rm{W}/4$ \\
    \hline
    \multirow{7}{*}{Alpha} & CONV + BN + RELU & (3 $\times$ 3), (1 $\times$ 1) & $e(x)$ & $256 \times \rm{H}/4 \times \rm{W}/4$ & A1 & $128 \times \rm{H}/4 \times \rm{W}/4$ \\
     & CONV + BN + RELU & (3 $\times$ 3), (1 $\times$ 1) & A1 & $128 \times \rm{H}/4 \times \rm{W}/4$ & A2 & $32 \times \rm{H}/4 \times \rm{W}/4$ \\
     & Upsampling & - & A2 & $32 \times \rm{H}/4 \times \rm{W}/4$ & A3 & $32 \times \rm{H} \times \rm{W}$ \\
     & CONV + BN + RELU & (3 $\times$ 3), (1 $\times$ 1)& $x$ & $3 \times \rm{H} \times \rm{W}$ & A4 & $32 \times \rm{H} \times \rm{W}$ \\
     & CONV + BN + RELU & (3 $\times$ 3), (1 $\times$ 1)& A3 and A4 & $64 \times \rm{H} \times \rm{W}$ & A5 & $32 \times \rm{H} \times \rm{W}$ \\
     & CONV + BN + RELU & (3 $\times$ 3), (1 $\times$ 1)& A5 & $32 \times \rm{H} \times \rm{W}$ & A6 & $32 \times \rm{H} \times \rm{W}$ \\
     & CONV + Sigmoid &(3 $\times$ 3), (1 $\times$ 1) & A6 & $32 \times \rm{H} \times \rm{W}$ & Alpha for decomposition & $3 \times \rm{H} \times \rm{W}$ \\
     \hline
    \multirow{2}{*}{Font} & CONV + BN + RELU&(3 $\times$ 3), (1 $\times$ 1) & $e^{t}_{b}(x)$ & $256 \times 1 \times 1$ & F1 & $128 \times 1 \times 1$ \\
     & CONV & (1 $\times$ 1), (1 $\times$ 1) & F1 & $128 \times 1 \times 1$ & Font categories & $100 \times 1 \times 1$ \\
     \hline
    \multirow{3}{*}{Effects visiblity} & CONV + BN + RELU& (3 $\times$ 3), (1 $\times$ 1) & $e^{t}_{b}(x)$ & $256 \times 1 \times 1 $ & V1 & $128 \times 1 \times 1$ \\
     & CONV & (1 $\times$ 1), (1 $\times$ 1)& V1 & $128 \times 1 \times 1$ & Shadow visibility & $2 \times 1 \times 1$ \\
     & CONV & (1 $\times$ 1), (1 $\times$ 1)& V1 & $128 \times 1 \times 1 $ & Border visibility & $2 \times 1 \times 1$ \\
     \hline
    \multirow{4}{*}{Effects params} & CONV + BN + RELU& (3 $\times$ 3), (1 $\times$ 1) & $e^{t}_{b}(x)$ & $256 \times 1 \times 1 $ & P1 & $128 \times 1 \times 1 $ \\
     & CONV + Tanh & (1 $\times$ 1), (1 $\times$ 1) & P1 & $128 \times 1 \times 1$ & Shadow offset & $2 \times 1 \times 1$ \\
     & CONV + Sigmoid & (1 $\times$ 1), (1 $\times$ 1) & P1 & $128 \times 1 \times 1$ & Shadow blur & $1 \times 1 \times 1$ \\
     & CONV & (1 $\times$ 1), (1 $\times$ 1)& P1 & $128 \times 1 \times 1$ & Border weights & $5 \times 1 \times 1$ \\
    \hline
\end{tabular}
}
\end{table*}

\end{document}